# Reproducible evaluation of classification methods in Alzheimer's disease: framework and application to MRI and PET data

Jorge Samper-González[a,b], Ninon Burgos[a,b], Simona Bottani[a,b], Sabrina Fontanella[a,b], Pascal Lu[a,b], Arnaud Marcoux[a,b], Alexandre Routier[a,b], Jérémy Guillon[a,b], Michael Bacci[a,b], Junhao Wen[a,b], Anne Bertrand[a,b,c,*], Hugo Bertin[d], Marie-Odile Habert[d,e], Stanley Durrleman[a,b], Theodoros Evgeniou[f], Olivier Colliot[a,b,c,g], for the Alzheimer's Disease Neuroimaging Initiative[1] and the Australian Imaging Biomarkers and Lifestyle flagship study of ageing[2]

[a] *Inria Paris, ARAMIS project-team, 75013 Paris, France*
[b] *Sorbonne Université, UPMC Univ Paris 06, Inserm, CNRS, Institut du cerveau et la moelle (ICM), AP-HP - Hôpital Pitié-Salpêtrière, 75013 Paris, France*
[c] *AP-HP, Department of Neuroradiology, Pitié-Salpêtrière Hospital, Paris, France*
[d] *Sorbonne Université, UPMC Univ Paris 06, Inserm, CNRS, LIB, AP-HP, 75013 Paris, France*
[e] *AP-HP, Department of Nuclear Medicine, Pitié-Salpêtrière Hospital, Paris, France*
[f] *INSEAD, Bd de Constance, 77305 Fontainebleau, France*
[g] *AP-HP, Department of Neurology, Pitié-Salpêtrière Hospital, Paris, France*
[*] *Deceased, March 2nd, 2018*

**Correspondence to:**
Olivier Colliot, PhD - olivier.colliot@upmc.fr
Jorge Samper-González - jorge.samper-gonzalez@inria.fr
ICM – Brain and Spinal Cord Institute
ARAMIS team
Pitié-Salpêtrière Hospital
47-83, boulevard de l'Hôpital, 75651 Paris Cedex 13, France

[1] Data used in preparation of this article were obtained from the Alzheimer's Disease Neuroimaging Initiative (ADNI) database (adni.loni.usc.edu). As such, the investigators within the ADNI contributed to the design and implementation of ADNI and/or provided data but did not participate in analysis or writing of this report. A complete listing of ADNI investigators can be found at:

http://adni.loni.usc.edu/wp-content/uploads/how_to_apply/ADNI_Acknowledgement_List.pdf

[2] Data used in the preparation of this article was obtained from the Australian Imaging Biomarkers and Lifestyle flagship study of ageing (AIBL) funded by the Commonwealth Scientific and Industrial Research Organisation (CSIRO) which was made available at the ADNI database (www.loni.usc.edu/ADNI). The AIBL researchers contributed data but did not participate in analysis or writing of this report. AIBL researchers are listed at www.aibl.csiro.au.

# Abstract

A large number of papers have introduced novel machine learning and feature extraction methods for automatic classification of Alzheimer's disease (AD). However, while the vast majority of these works use the public dataset ADNI for evaluation, they are difficult to reproduce because different key components of the validation are often not readily available. These components include selected participants and input data, image preprocessing and cross-validation procedures. The performance of the different approaches is also difficult to compare objectively. In particular, it is often difficult to assess which part of the method (e.g. preprocessing, feature extraction or classification algorithms) provides a real improvement, if any. In the present paper, we propose a framework for reproducible and objective classification experiments in AD using three publicly available datasets (ADNI, AIBL and OASIS). The framework comprises: i) automatic conversion of the three datasets into a standard format (BIDS); ii) a modular set of preprocessing pipelines, feature extraction and classification methods, together with an evaluation framework, that provide a baseline for benchmarking the different components. We demonstrate the use of the framework for a large-scale evaluation on 1960 participants using T1 MRI and FDG PET data. In this evaluation, we assess the influence of different modalities, preprocessing, feature types (regional or voxel-based features), classifiers, training set sizes and datasets. Performances were in line with the state-of-the-art. FDG PET outperformed T1 MRI for all classification tasks. No difference in performance was found for the use of different atlases, image smoothing, partial volume correction of FDG PET images, or feature type. Linear SVM and L2-logistic regression resulted in similar performance and both outperformed random forests. The classification performance increased along with the number of subjects used for training. Classifiers trained on ADNI generalized well to AIBL and OASIS. All the code of the framework and the experiments is publicly available: general-purpose tools have been integrated into the Clinica software (www.clinica.run) and the paper-specific code is available at: https://gitlab.icm-institute.org/aramislab/AD-ML.

**Keywords**:

classification, reproducibility, Alzheimer's disease, magnetic resonance imaging, positron emission tomography, open-source

# 1. Introduction

Alzheimer's disease (AD) affects over 20 million people worldwide. Identification of AD at an early stage is important for adequate care of patients and for testing of new treatments. Neuroimaging provides useful information to identify AD (Ewers et al., 2011): atrophy due to gray matter loss with anatomical magnetic resonance imaging (MRI), hypometabolism with $^{18}$F-fluorodeoxyglucose positron emission tomography (FDG PET), accumulation of amyloid-beta protein with amyloid PET imaging. A major interest is then to analyse those markers to identify AD at an early stage. In particular, machine learning methods have the potential to assist in identifying patients with AD by learning discriminative patterns from neuroimaging data.

A large number of machine learning approaches have been proposed to classify and predict AD stages (see (Falahati et al., 2014; Haller et al., 2011; Rathore et al., 2017) for reviews). Some of them make use of a single imaging modality (usually anatomical MRI) (Cuingnet et al., 2011; Fan et al., 2008; Klöppel et al., 2008; Liu et al., 2012; Tong et al., 2014) and others have proposed to combine multiple modalities (MRI and PET images, fluid biomarkers) (Gray et al., 2013; Jie et al., 2015; Teipel et al., 2015; Young et al., 2013; Yun et al., 2015; Zhang et al., 2011). Validation and comparison of such approaches require a large number of patients followed over time. A large number of published works uses the publicly available Alzheimer's Disease Neuroimaging Initiative (ADNI) dataset. However, the objective comparison between their results is almost impossible because they differ in terms of: i) subsets of patients (with unclear specification of selection criteria); ii) image preprocessing pipelines (and thus it is not clear if the superior performance comes from the classification or the preprocessing); iii) feature extraction and selection; iv) machine learning algorithms; v) cross-validation procedures and vi) reported evaluation metrics. Because of these differences, it is arduous to conclude which methods perform the best, and even whether a given modality provides useful additional information. As a result, the practical impact of these works has remained very limited. Moreover, the vast majority of these works use the ADNI dataset (ADNI1 for earlier papers and most often a combination of ADNI1, ADNI-GO and ADNI2 for more recent works). Therefore, assessment of generalization to another dataset is rarely done, even though other publicly available datasets exist such as the Australian Imaging Biomarker and Lifestyle study (AIBL) and the Open Access Series of Imaging Studies (OASIS).

Comparison papers (Cuingnet et al., 2011; Sabuncu et al., 2015) and challenges (Allen et al., 2016; Bron et al., 2015) have been an important step towards objective evaluation of machine learning methods by allowing the benchmark of different methods on the same dataset and with the same preprocessing. Nevertheless, such studies provide a "static" assessment of methods. Evaluation datasets are used in their current state at the time of the study, whereas new patients are continuously included in studies such as ADNI. Similarly, they are limited to the classification and preprocessing methods that were used at the time of the study. It is thus difficult to complement them with new approaches.

In this paper, we propose a framework for the reproducible evaluation of machine learning algorithms in AD and demonstrate its use on classification of PET and MRI data obtained from three publicly available datasets: ADNI, AIBL and OASIS. Specifically, our contributions are three-fold:

i) a framework for the management of publicly available datasets and their continuous update with new subjects, and in particular tools for fully automatic conversion into the Brain Imaging Data Structure[3] (BIDS) format (Gorgolewski et al., 2016);

ii) a modular set of preprocessing pipelines, feature extraction and classification methods, together with an evaluation framework, that provide a baseline for benchmarking of different components;

iii) a large-scale evaluation on T1 MRI and PET data from three publicly available neuroimaging datasets (ADNI, AIBL and OASIS).

We demonstrate the use of this framework for automatic classification from T1 MRI and PET data obtained from three datasets (ADNI, AIBL and OASIS). We assess the influence of various components on the classification performance: modality (T1 MRI or PET), feature type (voxel or regional features), preprocessing, diagnostic criteria (standard NINCDS/ADRDA criteria or amyloid-refined criteria), classification algorithm. Experiments were first performed on the ADNI, AIBL and OASIS datasets independently, and the generalization of the results was assessed by applying classifiers trained on ADNI to the AIBL and OASIS data.

All the code of the framework and the experiments is publicly available: general-purpose tools have been integrated into Clinica[4] (Routier et al., 2018), an open-source software platform that we developed to process data from neuroimaging studies, and the paper-specific code is available at: https://gitlab.icm-institute.org/aramislab/AD-ML.

---

[3] http://bids.neuroimaging.io
[4] http://clinica.run

## 2. Materials

### 2.1. Datasets

Part of the data used in the preparation of this article were obtained from the Alzheimer's Disease Neuroimaging Initiative database (adni.loni.usc.edu). The ADNI was launched in 2003 as a public-private partnership, led by Principal Investigator Michael W. Weiner, MD. The primary goal of ADNI has been to test whether serial MRI, PET, other biological markers, and clinical and neuropsychological assessment can be combined to measure the progression of mild cognitive impairment (MCI) and early AD. Over 1,650 participants were recruited across North America during the three phases of the study (ADNI1, ADNI GO and ADNI2). Around 400 participants were diagnosed with AD, 900 with MCI and 350 were control subjects. Three main criteria were used to classify the subjects (Petersen et al., 2010). The normal subjects had no memory complaints, while the subjects with MCI and AD both had to have complaints. CN and MCI subjects had a mini-mental state examination (MMSE) score between 24 and 30 (inclusive), and AD subjects between 20 and 26 (inclusive). The CN subjects had a clinical dementia rating (CDR) score of 0, the MCI subjects of 0.5 with a mandatory requirement of the memory box score being 0.5 or greater, and the AD subjects of 0.5 or 1. The other criteria can be found in (Petersen et al., 2010).

We also used data collected by the AIBL study group. Similarly to ADNI, the Australian Imaging, Biomarker & Lifestyle Flagship Study of Ageing seeks to discover which biomarkers, cognitive characteristics, and health and lifestyle factors determine the development of AD. AIBL has enrolled 1100 participants and collected over 4.5 years worth of longitudinal data: 211 AD patients, 133 MCI patients and 768 comparable healthy controls. AIBL study methodology has been reported previously (Ellis et al., 2010, 2009). Briefly, the MCI diagnoses were made according to a protocol based on the criteria of, (Winblad et al., 2004) and the AD diagnoses on the NINCDS-ADRDA criteria (McKhann et al., 1984). Note that about half of the subjects diagnosed as healthy controls reported memory complaints (Ellis et al., 2010, 2009).

Finally, we used data from the Open Access Series of Imaging Studies project whose aim is to make MRI datasets of the brain freely available to the scientific community. We focused on the "Cross-sectional MRI Data in Young, Middle Aged, Nondemented and Demented Older Adults" set (Marcus et al., 2007), which consists of a cross-sectional collection of 416 subjects aged 18 to 96. 100 of the included subjects over the age of 60 have been clinically diagnosed with very mild to moderate AD. The criteria used to evaluate the diagnosis was the CDR score. All participants with a CDR greater than 0 were diagnosed with probable AD. Note that there are no MCI subjects in OASIS.

### 2.2. Participants

#### 2.2.1. ADNI

Three subsets were created from the ADNI dataset: $ADNI_{T1w}$, $ADNI_{CLASS}$ and $ADNI_{CLASS, A\beta}$. $ADNI_{T1w}$ comprises all participants (N=1,628) for whom a T1-weighted (T1w) MR image was available at baseline. $ADNI_{CLASS}$ comprises 1,159 participants for whom a T1w MR image and an FDG PET scan, with a known effective resolution, were available at baseline. $\text{ADNI}_{\text{CLASS, A}\beta}$ is a subset of $\text{ADNI}_{\text{CLASS}}$ that comprises 918 participants with a known amyloid status determined from a PiB or an AV45 PET scan using 1.47 and 1.10 as cutoffs, respectively (Landau et al., 2013). For each ADNI subset, five diagnosis groups were considered:

- CN: subjects who were diagnosed as CN at baseline;

- AD: subjects who were diagnosed as AD at baseline;

- MCI: subjects who were diagnosed as MCI, EMCI or LMCI at baseline;

- pMCI: subjects who were diagnosed as MCI, EMCI or LMCI at baseline, were followed during at least 36 months and progressed to AD between their first visit and the visit at 36 months;

- sMCI: subjects who were diagnosed as MCI, EMCI or LMCI at baseline, were followed during at least 36 months and did not progress to AD between their first visit and the visit at 36 months.

Naturally, all participants in the pMCI and sMCI groups are also in the MCI group. Note that the reverse is false, as some MCI subjects did not convert to AD but were not followed long enough to state whether they were sMCI or pMCI. We did not consider the subjects with significant memory concerns (SMC) as this category only exists in ADNI 2.

Tables 1, 2 and 3 summarize the demographics, and the MMSE and global CDR scores of the participants composing $\text{ADNI}_{\text{T1w}}$, $\text{ADNI}_{\text{CLASS}}$ and $\text{ADNI}_{\text{CLASS, A}\beta}$.

**Table 1** Summary of participant demographics, mini-mental state examination (MMSE) and global clinical dementia rating (CDR) scores for $\text{ADNI}_{\text{T1w}}$.

|     | N   | Age | Gender | MMSE | CDR |
| --- | --- | --- | --- | --- | --- |
| CN  | 418 | 74.7 ± 5.8 [56.2, 89.6] | 209 M / 209 F | 29.1 ± 1.1 [24, 30] | 0: 417; 0.5: 1 |
| MCI | 868 | 73.0 ± 7.6 [54.4, 91.4] | 512 M / 356 F | 27.6 ± 1.8 [23, 30] | 0: 2; 0.5: 865; 1: 1 |
| AD  | 342 | 75.0 ± 7.8 [55.1, 90.9] | 189 M / 153 F | 23.2 ± 2.1 [18, 28] | 0.5: 165; 1: 176; 2: 1 |

*Values are presented as mean ± SD [range]. M: male, F: female*

**Table 2** Summary of participant demographics, mini-mental state examination (MMSE) and global clinical dementia rating (CDR) scores for ADNI$_{CLASS}$.

|  | N | Age | Gender | MMSE | CDR |
|---|---|---|---|---|---|
| CN | 282 | 74.3 ± 5.9 [56.2, 89.0] | 147 M / 135 F | 29.0 ± 1.2 [24, 30] | 0: 281; 0.5: 1 |
| MCI | 640 | 72.7 ± 7.5 [55.0, 91.4] | 378 M / 262 F | 27.8 ± 1.8 [23, 30] | 0: 1; 0.5: 638; 1:1 |
| sMCI | 342 | 71.8 ± 7.5 [55.0, 88.6] | 202 M / 140 F | 28.1 ± 1.6 [23, 30] | 0.5: 342 |
| pMCI | 167 | 74.9 ± 6.9 [55.0, 88.3] | 98 M / 69 F | 27.0 ± 1.7 [24, 30] | 0.5: 166; 1: 1 |
| AD | 237 | 74.9 ± 7.8 [55.1, 90.3] | 137 M / 100 F | 23.2 ± 2.1 [18, 27] | 0.5: 99; 1: 137; 2: 1 |

*Values are presented as mean ± SD [range]. M: male, F: female*

**Table 3** Summary of participant demographics, mini-mental state examination (MMSE) and global clinical dementia rating (CDR) scores for ADNICLASS, Aß. The amyloid status (Aß-: negative, Aß+: positive) was determined from each participant's amyloid PET scan (PiB or AV45).

|  |  | N | Age | Gender | MMSE | CDR |
|---|---|---|---|---|---|---|
| CN | Aß- | 116 | 72.2 ± 6.1 [56.2, 89.0] | 60 M / 56 F | 29.0 ± 1.3 [24,30]] | 0: 115; 0.5: 1 |
|  | Aß+ | 63 | 75.7 ± 5.8 [65.7, 85.6] | 26 M / 37 F | 28.9 ± 1.1 [24, 30] | 0: 63 |
| MCI | Aß- | 195 | 70.0 ± 7.9 [55.0, 91.4] | 107 M / 88 F | 28.5 ± 1.4 [24, 30] | 0: 1; 0.5: 193; 1: 1 |
|  | Aß+ | 253 | 73.0 ± 6.8 [55.0, 87.8] | 142 M / 111 F | 27.7 ± 1.8 [23, 30] | 0.5: 253 |
| sMCI | Aß- | 147 | 69.7 ± 7.7 [55.5, 91.4] | 82 M / 65 F | 28.5 ± 1.4 [25, 30] | 0.5: 147 |
|  | Aß+ | 118 | 72.5 ± 6.5 [55.0, 87.8] | 67 M / 51 F | 27.9 ± 1.7 [23, 30] | 0.5: 118 |
| pMCI | Aß- | 10 | 70.1 ± 6.7 [60.0, 81.6] | 5 M / 5 F | 27.6 ± 2.0 [24, 30] | 0.5: 9; 1: 1 |
|  | Aß+ | 84 | 73.2 ± 6.9 [55.0, 85.9] | 47 M / 37 F | 27.2 ± 1.8 [24, 30] | 0.5: 84 |
| AD | Aß- | 18 | 77.2 ± 8.1 [60.6, 90.3] | 16 M / 2 F | 23.4 ± 2.0 [20, 26] | 0.5: 7; 1: 11 |
|  | Aß+ | 126 | 74.1 ± 8.1 [55.1, 90.3] | 65 M / 61 F | 22.9 ± 2.1 [19, 26] | 0.5: 54; 1: 71; 2: 16 |

### 2.2.2. AIBL

The AIBL dataset considered in this work is composed of 608 participants for whom a T1-weighted MR image was available at baseline. The criteria used to create the diagnosis groups are identical to the ones used for ADNI. Table 4 summarizes the demographics, and the MMSE and global CDR scores of the AIBL participants.

**Table 4** Summary of participant demographics, mini-mental state examination (MMSE) and global clinical dementia rating (CDR) scores for AIBL.

|      | N   | Age                | Gender      | MMSE                | CDR                       |
|------|-----|--------------------|-------------|---------------------|---------------------------|
| CN   | 442 | 72.5 ± 6.2 [60, 92] | 191 M / 251 F | 28.7 ± 1.2 [25, 30] | 0: 415; 0.5: 26; 1: 1     |
| MCI  | 94  | 75.2 ± 7.0 [60, 96] | 50 M / 44 F  | 27.1 ± 2.1 [20, 30] | 0: 6; 0.5: 87; 1: 1       |
| sMCI | 21  | 75.8 ± 6.1 [64, 87] | 12 M / 9 F   | 27.9 ± 1.6 [25, 30] | 0.5: 21                   |
| pMCI | 16  | 78.0 ± 7.3 [63, 91] | 8 M / 8 F    | 26.9 ± 2.0 [22, 30] | 0.5: 16                   |
| AD   | 72  | 73.4 ± 7.9 [55, 93] | 30 M / 42 F  | 20.5 ± 5.6 [6, 29]  | 0.5: 31; 1: 32; 2: 7; 3: 2 |

*Values are presented as mean ± SD [range]. M: male, F: female*

### 2.2.3. OASIS

The OASIS dataset considered in this work is composed of 193 participants aged 61 years or more (minimum age of the participants diagnosed with AD). Table 5 summarizes the demographics, and the MMSE and global CDR scores of the OASIS participants.

**Table 5** Summary of participant demographics, mini-mental state examination (MMSE) and global clinical dementia rating (CDR) scores for OASIS.

|    | N   | Age                 | Gender      | MMSE                 | CDR                    |
|----|-----|---------------------|-------------|----------------------|------------------------|
| CN | 93  | 76.8 ± 8.4 [62, 94] | 25 M / 68 F | 28.9 ± 1.21 [25, 30] | 0: 93                  |
| AD | 100 | 76.8 ± 7.1 [62, 96] | 41 M / 59 F | 24.3 ± 4.15 [14, 30] | 0.5: 70; 1: 28; 2: 2   |

*Values are presented as mean ± SD [range]. M: male, F: female*

## 2.3. Imaging data

### 2.3.1. ADNI

#### 2.3.1.1. T1-weighted MRI

The acquisition protocols of the 3D T1w images can be found in (Jack et al., 2008) for ADNI 1 and (Jack et al., 2010a) for ADNI GO/2. The images can be downloaded as they were acquired or after having undergone several preprocessing correction steps, which include correction of image geometry distortion due to gradient non-linearity (gradwarp), correction of the image intensity non-uniformity that occurs when RF transmission is performed with a more uniform body coil while reception is performed with a less uniform head coil (B1 non-uniformity), and reduction of intensity non-uniformity due to the wave or the dielectric effect at 3 T or of residual intensity non-uniformity for 1.5 T scans (N3) (Jack et al., 2010a, 2008).

#### 2.3.1.2. PET

The ADNI FDG PET protocol consisted of a dynamic acquisition of six five-minute frames (ADNI 1) or four five-minute frames (ADNI GO/2), 30 to 60 minutes post-injection (Jagust et al., 2015, 2010). Images at different stages of preprocessing (frame averaging, spatial alignment, interpolation to a standard voxel size, and smoothing to a common resolution of 8 mm full width at half maximum) are available for download. Even though not used in the experiments, 11C-Pittsburgh compound B (PIB), for ADNI 1, and 18F-Florbetapir, also known as AV45, for ADNI 1/GO/2, were acquired to image the deposition of amyloid in the brain. The protocol consisted of a dynamic acquisition of four five-minute frames from 50 to 70 minutes post-injection (Jagust et al., 2015, 2010). As for the FDG PET, images at different stages of preprocessing are available for download.

### 2.3.2. AIBL

The T1w MR images used for the AIBL subjects were acquired using the ADNI 3D T1w sequence, with 1 × 1 mm in-plane resolution and 1.2 mm slice thickness, TR/TE/TI=2300/2.98/900, flip angle 9°, and field of view 240 × 256 and 160 slices (Ellis et al., 2010). Even though they were not used in the experiments, Florbetapir, PiB and Flutemetamol PET data were also acquired.

### 2.3.3. OASIS

For each OASIS subject, three or four T1w images, with 1 × 1 mm in-plane resolution and 1.25 mm slice thickness, TR/TE/TI=9.7/4.0/20, flip angle 10°, field of view 256 × 256 and 128 slices, were acquired on a 1.5 T scanner in a single imaging session (Marcus et al., 2007). For each subject, an average of the motion-corrected co-registered images resampled to 1 mm isotropic voxels, as well as spatially normalized images, are also available for download.

# 3. Methods

We developed a unified set of tools for data management, image preprocessing, feature extraction, classification, and evaluation. These tools have been integrated into Clinica[5] (Routier et al., 2018), an open-source software platform that we developed. Conversion tools allow an easy update of the datasets as new subjects become available. The different components were designed in a modular way: processing pipelines using Nipype (Gorgolewski et al., 2011), and classification and evaluation tools using the scikit-learn[6] library (Pedregosa et al., 2011). This allows the development and testing of other methods as replacement for a given step, and the objective measurement of the impact of each component on the results. A simple command line interface is provided and the code can also be used as a Python library.

## 3.1. Converting datasets to a standardized data structure

Even though public datasets are extremely valuable, an important difficulty with these studies lies in the organization of the clinical and imaging data. As an example, the ADNI and AIBL imaging data, in the state they are downloaded, do not rely on community standards for data organization and lack of a clear structure. Multiple image acquisitions exist for a given visit of a participant and the complementary image information is contained in numerous csv files, making the exploration of the database and subject selection very complicated. To organize the data, we selected the BIDS format (Gorgolewski et al., 2016), a community standard enabling the storage of multiple neuroimaging modalities. Being based on a file hierarchy rather than on a database management system, BIDS can be easily deployed in any environment. Very importantly, we provide the code that automatically performs the conversion of the data as they were downloaded to the BIDS organized version, for all the datasets used: ADNI, AIBL and OASIS. This allows direct reproducibility by other groups without having to redistribute the dataset, which is not allowed in the case of ADNI and AIBL. We also provide tools for subject selection according to desired imaging modalities, duration of follow up and diagnoses, which makes possible the use of the same groups with the largest possible number of subjects across studies. Finally, we propose a BIDS-inspired standardized structure for all the outputs of the experiments.

### 3.1.1. Conversion of the ADNI dataset to BIDS

The ADNI to BIDS converter requires the user to have downloaded all the ADNI study data (tabular data in csv format) and the imaging data of interest. Note that the downloaded files must be kept exactly as they were downloaded. The following steps are performed by the automatic converter (no user intervention is required). To convert the imaging data to BIDS, a list of subjects with their sessions is first obtained from the ADNIMERGE spreadsheet. This list is compared for each modality of interest to the list of scans available, as provided by modality-specific csv files (e.g. MRILIST.csv). If the modality was acquired for a specific pair of subject-session, and several scans and/or preprocessed images are available, only one is converted. Regarding the T1 scans, when several are available for a single session, the preferred scan (as identified in MAYOADIRL_MRI_IMAGEQC_12_08_15.csv) is chosen. If a preferred scan is not specified then the higher quality scan (as defined in MRIQUALITY.csv) is selected. If no quality control is found, then we choose the first scan. Gradwarp and B1-inhomogeneity corrected images are selected when available as these corrections can be performed in a clinical setting, otherwise the original image is selected. 1.5 T images are preferred for ADNI 1 since they are available for a larger number of patients. Regarding the FDG PET scans, the images co-registered and averaged across time frames are selected. The scans failing quality control (if specified in PETQC.csv) are discarded. Note that AV45 PET scans are also converted, though not used in the experiments. Once the images of interest have been selected and the paths to the image files identified, the imaging data can be converted to BIDS. When in dicom format, the images are first converted to nifti using the dcm2niix tool, or in case of error the dcm2nii tool (Li et al., 2016). The BIDS folder structure is generated by creating a subfolder for each of the subjects. A session folder is created inside each of the subject subfolders, and a modality folder is created inside each of the session subfolders. Finally, each image in nifti is copied to the appropriate folder and renamed to follow the BIDS specifications. Clinical data are also converted to BIDS. Data that do not change over time, such as the subject's sex, education level or diagnosis at baseline, are obtained from the ADNIMERGE spreadsheet and gathered in the participants.tsv file, located at the top of the BIDS folder hierarchy. The session-dependent data, such as the clinical scores, are obtained from specific csv files (e.g. MMSE.csv) and gathered in <subjectID>_session.tsv files in each participant subfolder. The clinical data being converted are defined in a spreadsheet (clinical_specifications_adni.xlsx) that is available with the code of the converter. The user can easily modify this file if he/she wants to convert additional clinical data.

---
[5] http://clinica.run

[6] http://scikit-learn.org

### 3.1.2. Conversion of the AIBL dataset to BIDS

The AIBL to BIDS converter requires the user to have downloaded the AIBL non-imaging data (tabular data in csv format) and the imaging data of interest. The conversion of the imaging data to BIDS relies on modality-specific csv files that provide the list of scans available. For each AIBL participant, the only T1w MR image available per session is converted. Note that even though they are not used in this work, we also convert the Florbetapir, PiB and Flutemetamol PET images (only one image per tracer is available for each session). Once the images of interest have been selected and the paths to the image files identified, the imaging data are converted to BIDS following the same steps as described in the above section. The conversion of the clinical data relies on the list of subjects and sessions obtained after the conversion of the imaging data and on the csv files containing the non-imaging data. Data that do not change over time are gathered in the participants.tsv file, located at the top of the BIDS folder hierarchy, while the session-dependent data are gathered in <subjectID>_session.tsv files in each participant subfolder. As for the ADNI converter, the clinical data being converted are defined in a spreadsheet (clinical_specifications.xlsx) available with the code of the converter, which the user can modify.

### 3.1.3. Conversion of the OASIS dataset to BIDS

The OASIS to BIDS converter requires the user to have downloaded the OASIS-1 imaging data and the associated csv file. To convert the imaging data to BIDS, the list of subjects is obtained from the downloaded folders. For each subject, among the multiple T1w MR images available, we select the average of the motion-corrected co-registered individual images resampled to 1 mm isotropic voxels, located in the SUBJ_111 subfolder. Once the paths to the image files have been identified, the images in Analyse format are converted to nifti using the mri_convert tool of FreeSurfer (Fischl, 2012), the BIDS folder hierarchy is created, and the images are copied to the appropriate folder and renamed. The clinical data are converted using the list of subjects obtained after the conversion of the imaging data and the csv file containing the non-imaging data, as described in the previous section.

## 3.2. Preprocessing pipelines

Two pipelines were developed to preprocess the anatomical T1w MRI and PET images. These pipelines have a modular structure based on Nipype allowing the user to easily connect and/or replace components, and rely on well established procedures using publicly available standard image processing tools. These pipelines are available in Clinica under the names t1-volume-* and pet-volume.

### 3.2.1. Preprocessing of T1-weighted MR images

For anatomical T1w MRI, the preprocessing pipeline was based on SPM12[7]. First, the Unified Segmentation procedure (Ashburner and Friston, 2005) is used to simultaneously perform tissue segmentation, bias correction and spatial normalization of the input image. Next, a group template is created using DARTEL, an algorithm for diffeomorphic image registration (Ashburner, 2007), from the subjects' tissue probability maps on the native space, usually GM, WM and CSF tissues, obtained at the previous step. Here, not only the group template is obtained, but also the deformation fields from each subject's native space into the DARTEL template space. Lastly, the DARTEL to MNI method (Ashburner, 2007) is applied, providing a registration of the native space images into the MNI space: for a given subject its flow field into the DARTEL template is combined with the transformation of the DARTEL template into MNI space, and the resulting transformation is applied to the subject's different tissue maps. As a result, all the images are in a common space, providing a voxel-wise correspondence across subjects.

### 3.2.2. Preprocessing of PET images

The PET preprocessing pipeline relies on SPM12 and on the PETPVC[8] tool for partial volume correction (PVC) (Thomas et al., 2016). We assume that each PET image has a corresponding T1w image that has been preprocessed using the pipeline described above. The first step is to perform a registration of the PET image to the corresponding T1w image in native space using the Co-register method of SPM (Friston et al., 1995). An optional PVC step with the regional voxel-based (RBV) method (Thomas et al., 2011) can be performed using as input regions the different tissue maps from the T1w in native space. Then, the PET image is registered into MNI space using the same transformation as for the corresponding T1w (the DARTEL to MNI method is used). The PET image in MNI space is then intensity normalized according to a reference region (eroded pons for FDG PET) and we obtain a standardized uptake value ratio (SUVR) map. Finally, we mask the non-brain regions using a binary mask resulting from thresholding the sum of the GM, WM and CSF tissue probability maps for the subject in MNI space. The resulting masked SUVR images are also in a common space and provide voxel-wise correspondence across subjects.

---

[7] http://www.fil.ion.ucl.ac.uk/spm/software/spm12/

[8] https://github.com/UCL/PETPVC

### 3.3. Feature extraction

Two types of features were extracted from the imaging data: voxel and region features. After preprocessing, both the T1w MRI and FDG PET images are in the MNI space. The first type of features simply corresponds, for each image, to all the voxels in the brain. The signal obtained from the T1w MR images is the gray matter density and the one obtained from the FDG PET images is the SUVR.

Regional features correspond to the average signal (gray matter density or SUVR, respectively) computed in a set of regions of interest (ROIs) obtained from different atlases, also in MNI space. The five atlases selected contain both cortical and subcortical regions, and cover the brain areas affected by AD. They are described below:

- AAL2 (Tzourio-Mazoyer et al., 2002) is an anatomical atlas based on a single subject. It is the updated version of AAL, which is probably the most widely used parcellation map in the neuroimaging literature. It was built using manual tracing on the spatially normalized single-subject high-resolution T1 volume in MNI space (Holmes et al., 1998). It is composed of 120 regions covering the whole cortex as well as the main subcortical structures.

- AICHA (Joliot et al., 2015) is a functional atlas based on multiple subjects. It was built using parcellation of group-level functional connectivity profiles computed from resting-state fMRI data of 281 healthy subjects. It is composed of 345 regions covering the whole cortex as well as the main subcortical structures.

- Hammers (Gousias et al., 2008; Hammers et al., 2003) is an anatomical atlas based on multiple subjects. It was built using manual tracing on anatomical MRI from 30 healthy subjects. The individual subjects parcellations were then registered to MNI space to generate a probabilistic atlas as well as a maximum probability map. The latter was used in the present work. It is composed of 69 regions covering the whole cortex as well as the main subcortical structures.

- LPBA40 (Shattuck et al., 2008) is an anatomical atlas based on multiple subjects. It was built using manual tracing on anatomical MRI from 40 healthy subjects. The individual subject parcellations were then registered to MNI space to generate a maximum probability map. It is composed of 56 regions covering the whole cortex as well as the main subcortical structures.

- Neuromorphometrics[9] is an anatomical atlas based on multiple subjects. It was built using manual tracing on anatomical MRI from 30 healthy subjects. The individual subject parcellations were then registered to MNI space to generate a maximum probability map.

---

[9] www.neuromorphometrics.com

It is composed of 140 regions covering the whole cortex as well as the main subcortical structures. Data were made available for the "MICCAI 2012 Grand Challenge and Workshop on Multi-Atlas Labeling".

The main difference between the LBPA40, Hammers and Neuromorphometrics atlases is the degree of detail (i.e. the number of regions) of the anatomical parcellation.

## 3.4. Classification models

We considered three different classifiers: linear SVM, logistic regression with L2 regularization, and random forest, all available in Clinica. The linear SVM was used with both the voxel and the regional features because its computational complexity depends only on the number of subjects when using its dual form. On the other hand, the logistic regression with L2 regularization and random forest models were only used for the region-based analyses given that their complexity depends on the number of features, which becomes infeasible with images containing about 1 million voxels. We used the implementations of the scikit-learn library (Pedregosa et al., 2011).

For each of the tasks performed, we obtain the feature weights that describe the importance of a given feature for the current classification task. These weights are stored as part of the output of the classifications, as is the information to reconstruct the classifiers, like the optimal parameters found. We can obtain, for each classification, an image with the representation of weights across brain voxels or regions.

### 3.4.1. Linear SVM

The first method included is linear SVM. To reduce computational load, the Gram matrix $K = (k(x_i, x_j))_{i,j}$ was precalculated using a linear kernel $k$ for each pair of images $(x_i, x_j)$ (using the region or voxel features) for the provided subjects. This Gram matrix is used as input for the generic SVM. We chose to optimize the penalty parameter C of the error term. An advantage of SVM is that, when using a precomputed Gram matrix (dual SVM), computing time depends on the number of subjects, and not on the number of features. Given its simplicity, linear SVM is useful as a baseline to compare the performance of the different methods.

### 3.4.2. Logistic regression with L2 regularization

The second method is logistic regression with L2 regularization (which is classically used to reduce overfitting). We optimized, as for the linear SVM, the penalty parameter C of the error term. Logistic regression with L2 regularization directly optimizes the weights for each feature, and the number of features influences the training time. This is the reason why we only used it for regional features.

### 3.4.3. Random forest

The third classifier used is the random forest. Unlike both linear SVM and logistic regression, random forest is an ensemble method that fits a number of decision trees on various sub-samples of the dataset. The combined estimator prevents overfitting and improves the predictive accuracy. Based on the implementation provided by the scikit-learn library (Pedregosa et al., 2011), there is a large number of parameters that can be optimized. After preliminary experiments to assess which had a larger influence, we selected the following two hyperparameters to optimize: i) the number of trees in the forest; ii) the number of features to consider when looking for the best split. Random forest was only used for regional features and not voxel features, due to its high computational cost.

## 3.5. Evaluation strategy

### 3.5.1. Cross-validation

Evaluation of classification performances mainly followed the recent guidelines provided by (Varoquaux et al., 2017). Cross-validation (CV), the classical strategy to maintain the independence of the train set (used for fitting the model) and the test set (used to evaluate the performances), was performed. The CV procedure included two nested loops: an outer loop evaluating the classification performances and an inner loop used to optimize the hyperparameters of the model (C for SVM and L2 logistic regression, the number of trees and features for a split for the random forest). It should be noted that the use of an inner loop of CV is important to avoid biasing performances upward when optimizing hyperparameters. This step has not always been appropriately performed in the literature (Querbes et al., 2009; Wolz et al., 2011) leading to over-optimistic results, as presented in (Eskildsen et al., 2013; Maggipinto et al., 2017).

We implemented in Clinica three different outer CV methods: k-fold, repeated k-fold and repeated random splits (all of them stratified), using scikit-learn based tools (Pedregosa et al., 2011). The choice of the method would depend on the computational resources at hand. However, whenever possible, it is recommended to use repeated random splits with a large number of repetitions to yield more stable estimates of performances. Therefore, we used for each experiment 250 iterations of random splits. We report the full distribution of the evaluation metrics in addition to the mean and empirical standard-deviation, as done in (Raamana and Strother, 2017) that uses of neuropredict (Raamana, 2017). It should nevertheless be noted that there is no unbiased estimate of variance for cross-validation (Bengio and Grandvalet, 2004; Nadeau and Bengio, 2003) and that the empirical variance largely underestimates the true variance. This should be kept in mind when interpreting the empirical variance values. Also, we chose not to perform statistical testing of the performance of different classifiers. This is a complex matter for which there is no universal solution. In many publications, a standard t-test on cross-validation results is used. However, such an approach is way too liberal and should not be applied, as shown by Nadeau and Bengio (2003). Better behaved approaches have been proposed such as the conservative Z or the corrected resampled t-test (Nadeau and Bengio, 2003). However, such approaches must be used with caution because their behaviour depends on the data and the cross-validation set-up. We thus chose to avoid the use of statistical tests in the present paper, in order not to mislead the reader. Instead, we reported the full distributions of the metrics.

For hyperparameter optimization, we implemented an inner k-fold. For each split, the model with the highest balanced accuracy is selected, and then these selected models are averaged across splits to profit of model averaging, that should have a stabilizing effect. In the present paper, experiments were performed with k=10 for the inner loop.

### 3.5.2. Metrics

As output of the classification, we report the balanced accuracy, area under the ROC curve (AUC), accuracy, sensitivity, specificity and, in addition, the predicted class for each subject, so the user can calculate other desired metrics with this information.

## 3.6. Classification experiments

The different classification tasks considered in our analyses for each dataset, driven by the data availability, are detailed in Table 6. Details regarding the group compositions can be found in Tables 2, 3, 4 and 5. In general, we perform clinical diagnosis classification tasks, or "predictive" tasks of the evolution of MCI subjects. Note that tasks involving progression from MCI to AD were not performed for AIBL due to the small number of participants in the sMCI and pMCI categories. However, the framework would allow performing these experiments very easily when more progressive MCI subjects become publicly available in AIBL.

Depending on the type of features, the performance of several classifiers with different parameters was tested. For voxel features, the only classifier was the linear SVM. Four different levels of smoothing were applied to the images using a Gaussian kernel, from no smoothing to up to 12 mm full width at half maximum (FWHM). For region-based classification experiments, three classifiers were tested: linear SVM, logistic regression and random forest. The features were extracted using five atlases: AAL2, AICHA, Hammers, LPBA40 and Neuromorphometrics. This information is summarized in Table 7.

For the datasets under study, different imaging modalities were available: while both T1w MRI and FDG PET images were available for the ADNI participants, only T1w MRI were available for AIBL and OASIS participants. For each modality considered, both voxel and region features were extracted using the different parameters detailed in Table 7. All the classification experiments tested in this work are summarized in Table 8. If not otherwise stated, the FDG PET features were extracted from images that did not undergo PVC.

**Table 6** List of classification tasks for each dataset.

| tasks_ADNI | tasks_AIBL | tasks_OASIS |
|---|---|---|
| CN vs AD | CN vs AD | CN vs AD |
| CN vs pMCI | CN vs MCI | |
| sMCI vs pMCI | | |
| CN vs MCI | | |
| CN- vs AD+ | | |
| CN- vs pMCI+ | | |
| sMCI- vs pMCI+ | | |
| sMCI+ vs pMCI+ | | |
| MCI- vs MCI+ | | |

**Table 7** Summary of classifiers and parameters used for each type of features.

| | | |
|---|---|---|
| Voxel-based | Linear SVM | Smoothing 0 mm |
| | | Smoothing 4 mm |
| | | Smoothing 8 mm |
| | | Smoothing 12 mm |
| Region-based | Linear SVM | Atlas AAL2 |
| | | Atlas Neuromorphometrics |
| | | Atlas Hammers |
| | | Atlas LPBA40 |
| | | Atlas AICHA |
| | Logistic Regression | Atlas AAL2 |
| | | Atlas Neuromorphometrics |
| | | Atlas Hammers |
| | | Atlas LPBA40 |
| | | Atlas AICHA |
| | Random Forest | Atlas AAL2 |
| | | Atlas Neuromorphometrics |
| | | Atlas Hammers |
| | | Atlas LPBA40 |
| | | Atlas AICHA |

**Table 8** Summary of all the classification experiments run in our analysis for each dataset, imaging modality, feature type (different parameters tested, see Table 7) and task (more details in Table 6).

| Dataset | Imaging Modality | | Feature Type | Tasks |
|---|---|---|---|---|
| ADNI | T1w MRI | | Voxel-based | tasks_ADNI |
| | | | Region-based | tasks_ADNI |
| | FDG PET | With PVC | Voxel-based | tasks_ADNI |
| | | | Region-based | tasks_ADNI |
| | | Without PVC | Voxel-based | tasks_ADNI |
| | | | Region-based | tasks_ADNI |
| AIBL | T1w MRI | | Voxel-based | tasks_AIBL |
| | | | Region-based | tasks_AIBL |
| OASIS | T1w MRI | | Voxel-based | tasks_OASIS |
| | | | Region-based | tasks_OASIS |

# 4. Results

Here, we present a selection of the results that we believe are the most valuable. The complete results of all experiments (including other tasks, preprocessing parameters, features or classifiers) are available in the Supplementary Material as well as in the repository containing all the code and experiments (https://gitlab.icm-institute.org/aramislab/AD-ML). In the following subsections, we present the results using the balanced accuracy as performance metric but all the other metrics are available in the results.

## 4.1. Influence of the atlas

To assess the impact of the choice of atlas on the classification accuracy and to potentially identify a preferred atlas, the linear SVM classifier using regional features was selected. Features from T1w MRI and FDG PET images of ADNI participants were extracted using five different atlases: AAL2, AICHA, Hammers, LPBA40 and Neuromorphometrics. Three classification tasks were studied: CN vs AD, CN vs pMCI and sMCI vs pMCI.

As shown in Figure 1, no specific atlas provides the highest classification accuracy for all the tasks. For example, Neuromorphometrics and AICHA provide better results for CN vs AD on T1w and FDG PET images, along with LBPA40 for T1w, while AAL2 provides the highest balanced accuracy for CN vs pMCI and sMCI vs pMCI on both imaging modalities. The same analysis was performed on AIBL subjects (T1w MR images only) and, similarly, no atlas consistently performed better than others across tasks. For the following region-based experiments, the AAL2 atlas was chosen as reference atlas as it leads to good classification accuracies and is widely used in the neuroimaging community. Again, all other results are available in the repository.

## 4.2. Influence of the smoothing

T1w MRI and FDG PET images were not smoothed or smoothed using Gaussian kernels with FWHMs of 4 mm, 8 mm and 12 mm. To determine the influence of different smoothing degrees on the classification accuracy, a linear SVM classifier using voxel features was chosen. Three classification tasks were studied: CN vs AD, CN vs pMCI and sMCI vs pMCI. The results in Figure 2 show that, for most classification tasks, the balanced accuracy does not vary to a great extent with the smoothing kernel size. The only variations are observed for the CN vs pMCI and sMCI vs pMCI tasks when the features were extracted from T1w MR images: the balanced accuracy increases slightly with the kernel size. The same analysis was run using T1w MR images from the AIBL dataset. The mean balanced accuracy also increased slightly with the kernel size, but the standard deviations of the balanced accuracies are larger than for ADNI. As the degree of smoothing does not have a clear impact on the classification performance, we chose to present the subsequent results related to the voxel-based classification with a reference smoothing of 4 mm.

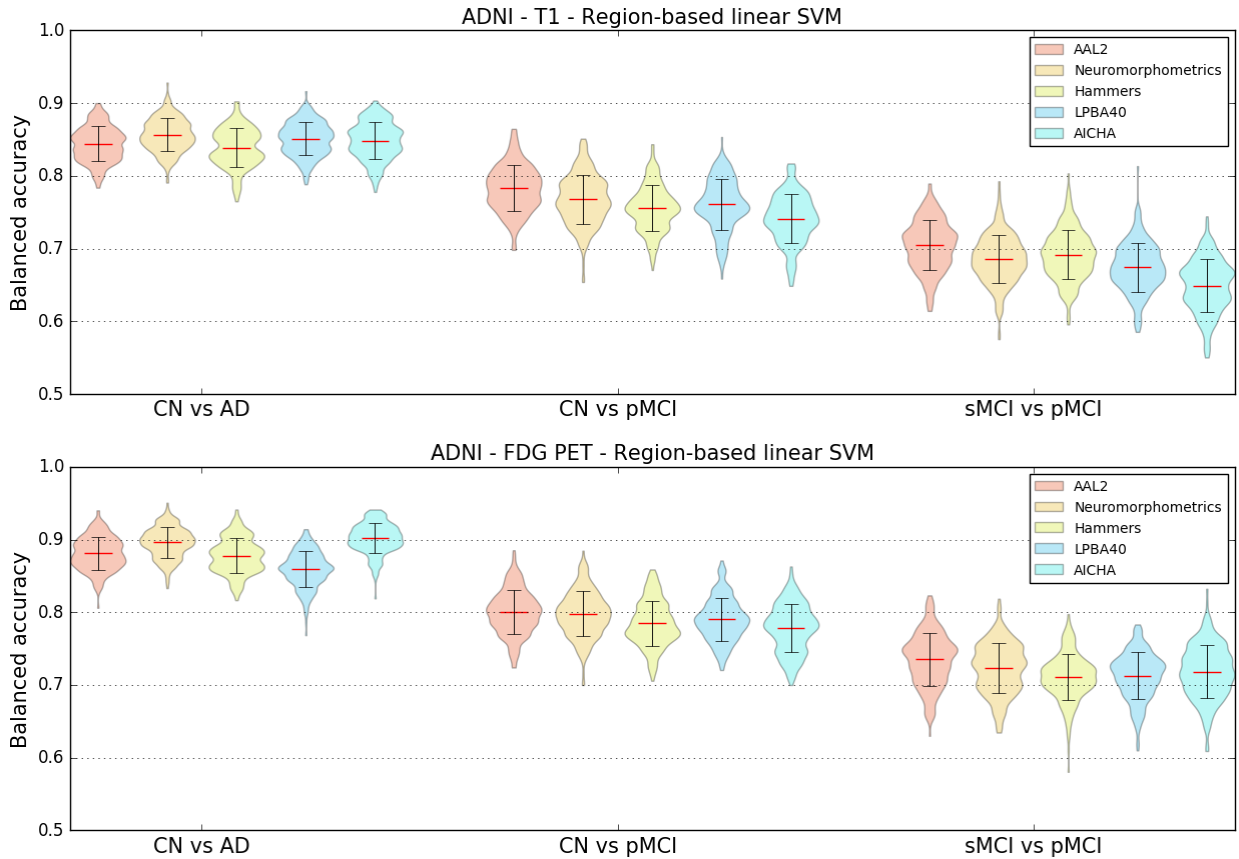

**Figure 1. Influence of atlas.** Distribution of the balanced accuracies obtained from the T1w MRI (top) and FDG PET (bottom) images of ADNI participants using the reference classifier (linear SVM) and regional features from different atlases for the CN vs AD, CN vs pMCI and sMCI vs pMCI tasks.

## 4.3. Influence of the type of features

We compared the balanced accuracies obtained for the voxel features with reference smoothing (Gaussian kernel of 4 mm FWHM) to the ones obtained for the regional features with reference atlas (AAL2) when using linear SVM classifiers. These features were extracted from T1w MRI and FDG PET images of ADNI participants. The same three classification tasks as before were evaluated.

The results, displayed in Table 9, do not show notable differences between the mean balanced accuracies obtained using voxel or regional features. In the case of the AIBL dataset, the balanced accuracy is higher for the region-based classification (for AD vs CN: voxel-based 0.79 [±0.059], region-based 0.86 [±0.042]), but we can observe that the corresponding standard deviations are high.

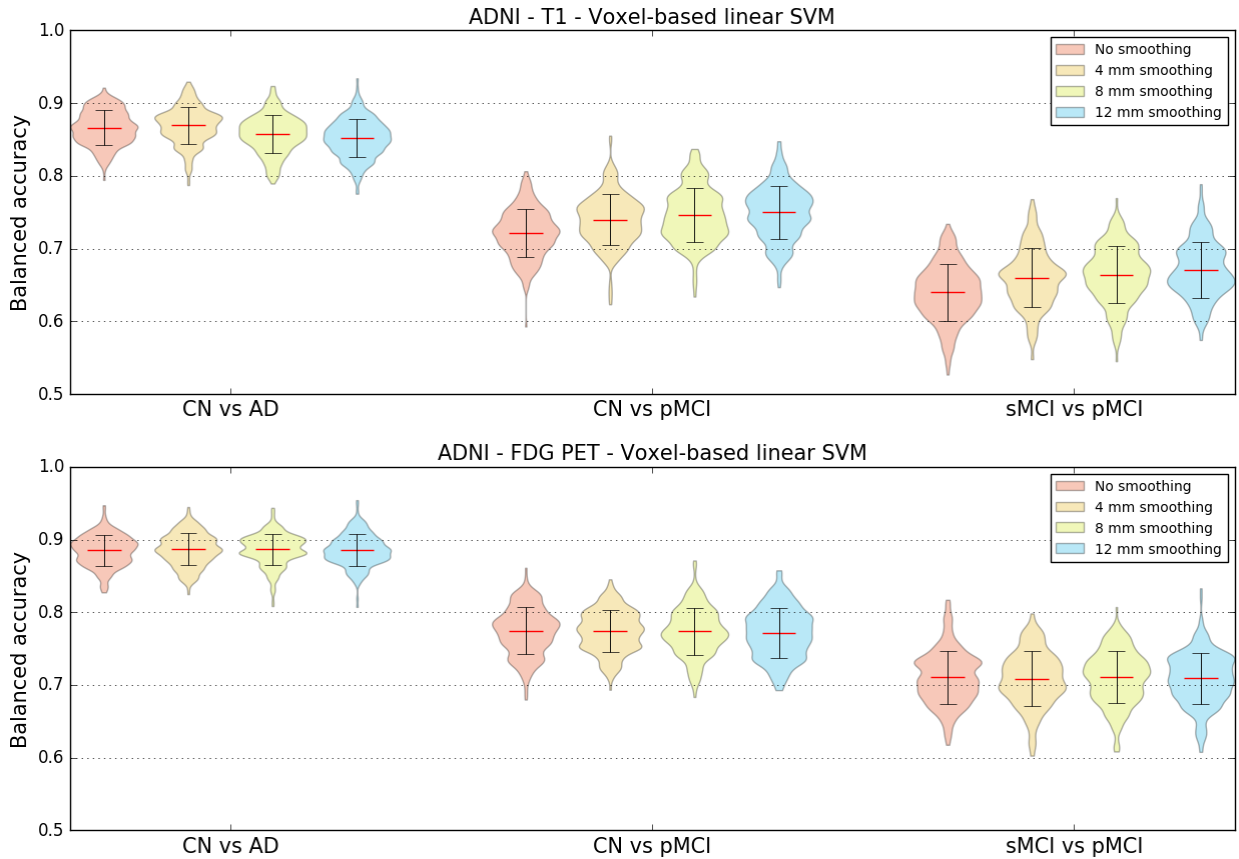

**Figure 2. Influence of smoothing.** Distribution of the balanced accuracy obtained from the T1w (top) and FDG PET (bottom) images of ADNI participants using the reference classifier (linear SVM) and voxel features with different degrees of smoothing for the CN vs AD, CN vs pMCI and sMCI vs pMCI tasks.

**Table 9. Influence of feature types.** Mean balanced accuracy and standard deviation obtained for three tasks (CN vs AD, CN vs pMCI and sMCI vs pMCI) using the reference classifier (linear SVM) with voxel (reference smoothing: 4 mm) and region (reference atlas: AAL2) features extracted from T1w MRI and FDG PET images of ADNI subjects.

|  | T1w – Linear SVM | | FDG PET – Linear SVM | |
| --- | --- | --- | --- | --- |
|  | Voxel-based (4 mm smoothing) | Region-based (AAL2 atlas) | Voxel-based (4 mm smoothing) | Region-based (AAL2 atlas) |
| CN vs AD | 0.87 ± 0.026 | 0.84 ± 0.024 | 0.88 ± 0.022 | 0.88 ± 0.023 |
| CN vs pMCI | 0.74 ± 0.035 | 0.78 ± 0.031 | 0.77 ± 0.028 | 0.80 ± 0.030 |
| sMCI vs pMCI | 0.66 ± 0.040 | 0.70 ± 0.034 | 0.71 ± 0.037 | 0.73 ± 0.036 |

## 4.4. Influence of the classification method

Region-based experiments were carried out using three different classifiers to evaluate if there were variations in balanced accuracies depending on the chosen classifier. Regional features were extracted using the reference AAL2 atlas from T1w MRI and FDG PET images of ADNI participants. The three previously defined classification tasks were performed.

The results displayed in Figure 3 show that both the linear SVM and logistic regression with L2 regularization models lead to similar balanced accuracies, consistently higher than the one obtained with random forest for all the tasks and imaging modalities tested.

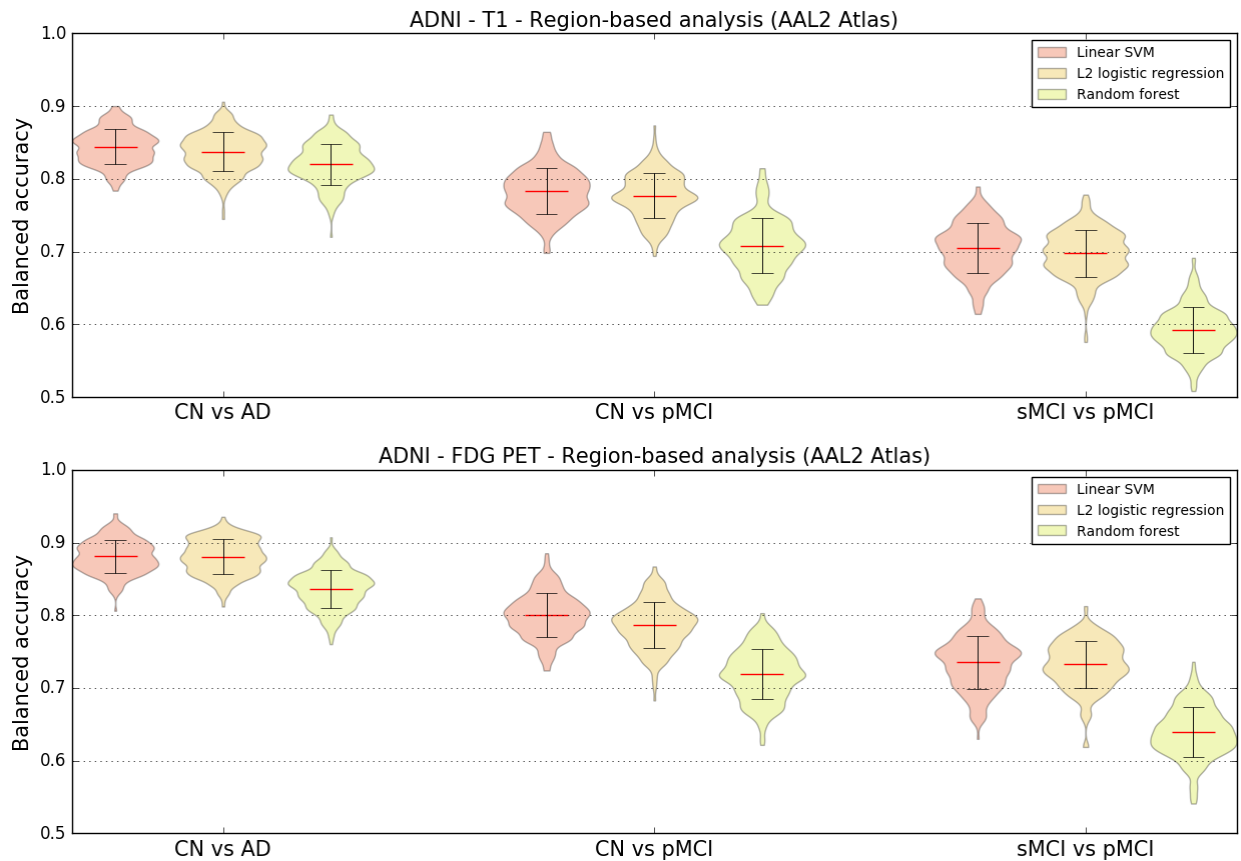

**Figure 3. Influence of classification method.** Distribution of the balanced accuracy obtained from the T1w MRI (top) and FDG PET (bottom) images of ADNI participants using different region-based classifiers (reference atlas: AAL2) for the CN vs AD, CN vs pMCI and sMCI vs pMCI tasks.

## 4.5. Influence of the partial volume correction of PET images

Both region and voxel-based analyses were performed using linear SVM classifiers to evaluate if correcting PET images for partial volume effect had an influence on the classification accuracy. FDG PET images of ADNI participants with and without PVC were used for these experiments.

The results displayed in Figure 4 show little difference between the balanced accuracies obtained with and without PVC. When using voxel features, the average balanced accuracy is almost identical no matter the presence or absence of PVC. Using regional features, there is a very small increase in mean balanced accuracy when the FDG PET images are not corrected for partial volume effect.

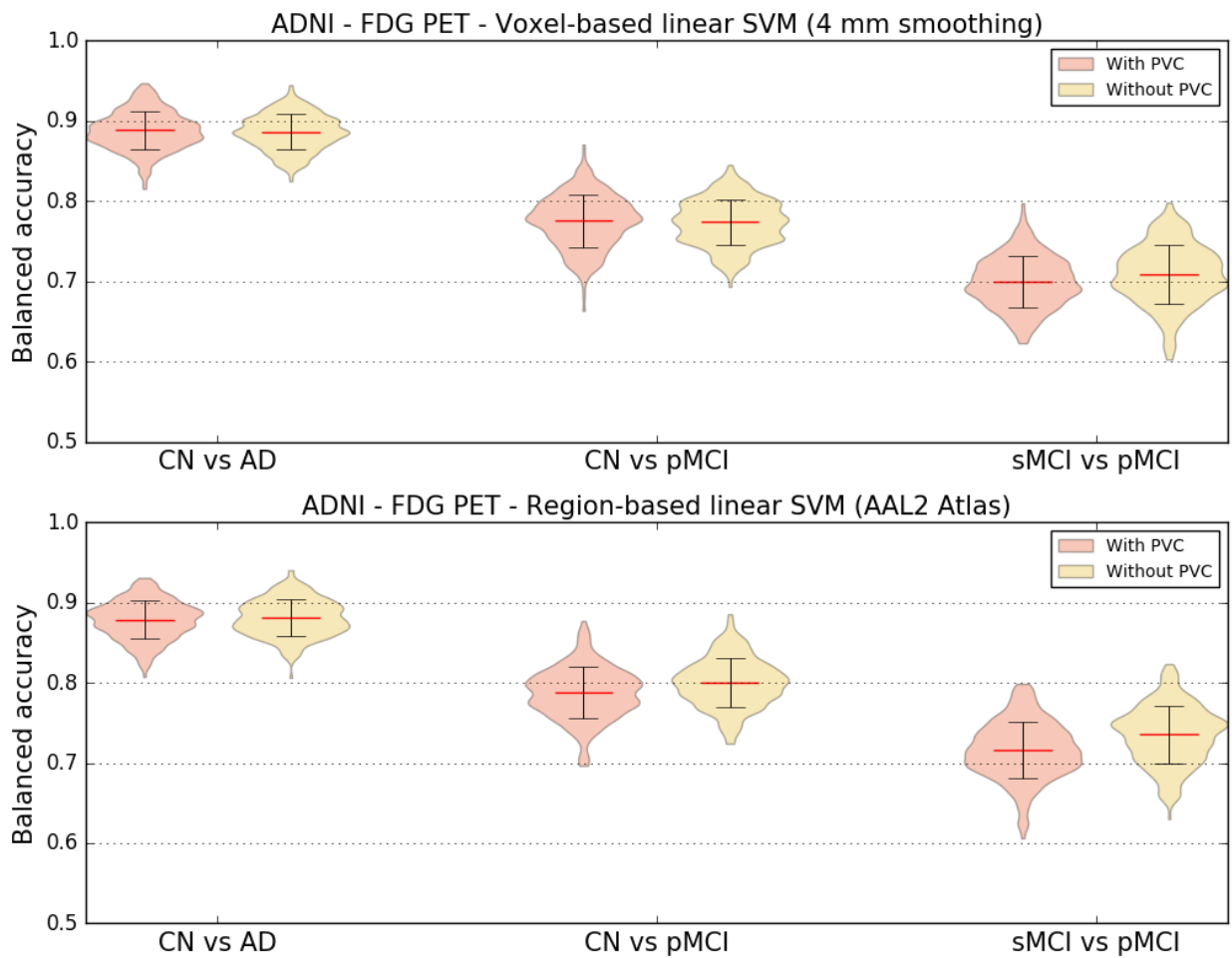

**Figure 4. Influence of partial volume correction.** Distribution of the balanced accuracy obtained from the FDG PET images of ADNI participants with and without PVC using the reference classifier (linear SVM) and regional features derived from the AAL2 atlas (top) and voxel features with 4 mm of smoothing (bottom) for the CN vs AD, CN vs pMCI and sMCI vs pMCI tasks.

## 4.6. Influence of the magnetic field strength

Most T1w scans of ADNI1 participants were acquired on 1.5 T scanners while 3 T scanners were used to acquire MR images for participants of ADNIGO/2. To assess whether the difference in field strength had an impact on the classification performance, we computed the balanced accuracy separately for the subjects who had 1.5 T and 3 T scans. The results are displayed on Table 10. We observed that, no matter the experiment, the balanced accuracy is always higher for the 3 T scan subset compared to the 1.5 T scan subset, which is not surprising as 3 T images should have a better signal-to-noise ratio.

**Table 10. Influence of magnetic field strength.** Mean balanced accuracy obtained for three tasks (CN vs AD, CN vs pMCI and sMCI vs pMCI) using the reference classifier (linear SVM) with voxel (reference smoothing: 4 mm) and region (reference atlas: AAL2) features extracted from T1w MR images of ADNI subjects. The mean balanced accuracy was computed separately for subjects whose images were acquired on 1.5 T (most ADNI1 subjects) and 3 T (ADNIGO/2 subjects) MRI scanners.

|  | Voxel-based (4 mm smoothing) | | Region-based (AAL2 atlas) | |
| --- | --- | --- | --- | --- |
|  | 1.5 T | 3 T | 1.5 T | 3 T |
| CN vs AD | 0.85 | 0.88 | 0.84 | 0.85 |
| CN vs pMCI | 0.73 | 0.74 | 0.77 | 0.78 |
| sMCI vs pMCI | 0.60 | 0.66 | 0.62 | 0.71 |

## 4.7. Influence of class imbalance

The tasks that we performed are done with unbalanced classes. Such class imbalance ranges from very mild (1.2 times more CN than AD for ADNI) to moderate (1.7 times more CN than pMCI and 2 times more sMCI than pMCI for ADNI) to very strong (6.1 times more CN than AD in AIBL). We aimed to assess if such class imbalance influenced the performance. To that purpose, we randomly sampled subgroups and performed experiments with 237 CN vs 237 AD, 167 pMCI vs 167 CN and 167 pMCI vs 167 pMCI for ADNI and 72 CN and 72 AD for AIBL. We ensured that the demographic and clinical characteristics of the balanced subsets did not differ from the original ones. Results are presented on **Figure 5**. For ADNI, the performance was similar to that obtained with the full population. For AIBL, the performance was substantially higher with balanced groups for the voxel-based features. It thus seems that a very strong class imbalance (as in the case of AIBL where the proportion is 6 to 1) leads to lower performance but that moderate class imbalance (up to 2 to 1 in ADNI) are adequately handled.

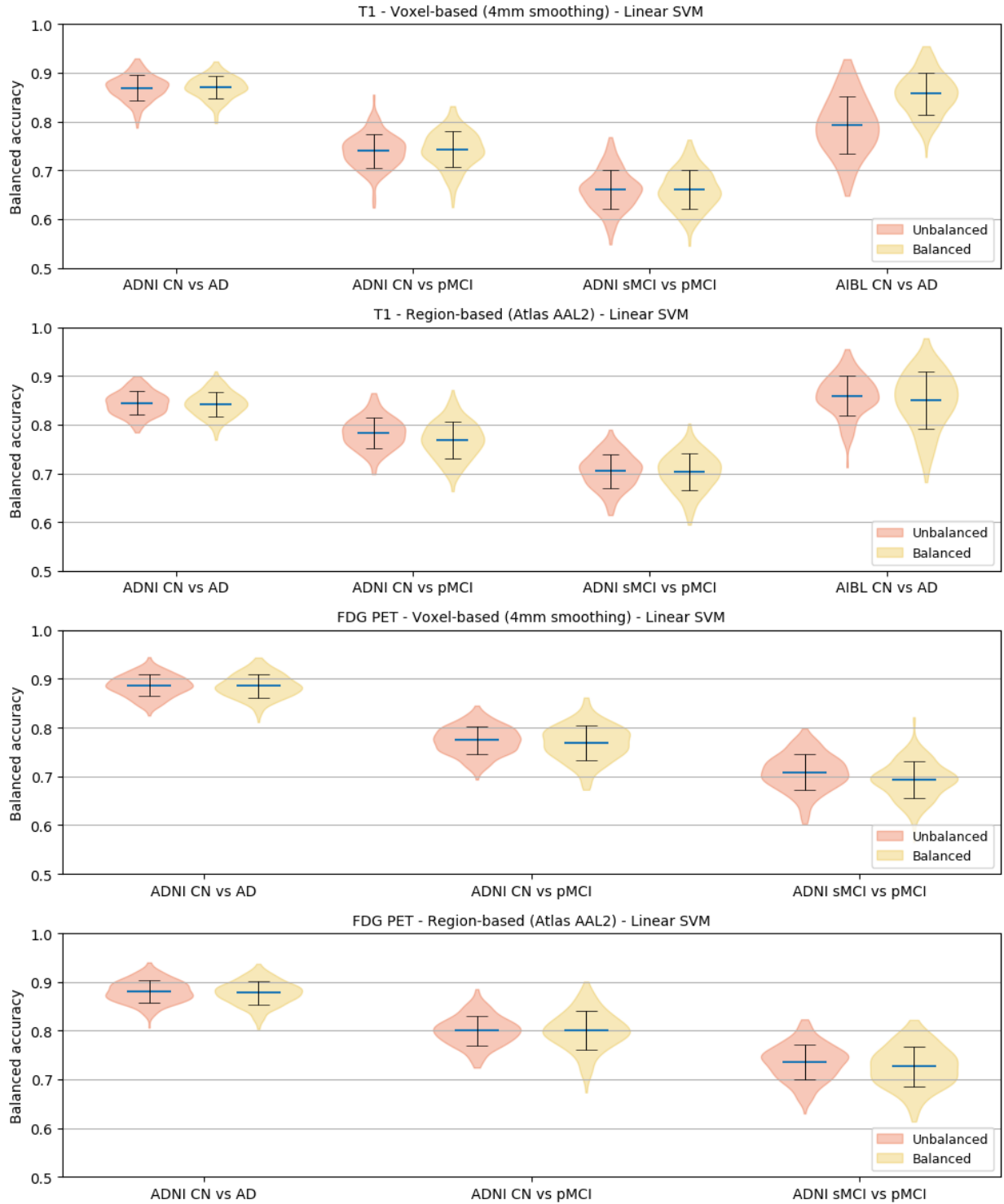

**Figure 5. Influence of class imbalance.** Distribution of the balanced accuracies obtained using voxel (reference smoothing: 4 mm) and regional (reference atlas: AAL2) features extracted from T1w MRI and FDG PET images using the reference classifier (linear SVM) when training using unbalanced and balanced datasets. Four tasks were tested: CN vs AD, CN vs pMCI and sMCI vs pMCI for ADNI subjects, and CN vs AD for AIBL subjects.

## 4.8. Influence of the dataset

We also wanted to know how consistent were the results across datasets, and thus we compared the classification performances obtained from ADNI, AIBL and OASIS, for the task of differentiating control subjects from patients with Alzheimer's disease. Voxel (4 mm smoothing) and regional (AAL2 atlas) features were extracted from T1w MR images and used with linear SVM classifiers. We tested two configurations: training and testing the classifiers on the same dataset, and training a classifier on ADNI and testing it on AIBL and OASIS.

Results are displayed in Table 11. Performances obtained on ADNI and AIBL were comparable and much higher than those obtained on OASIS. When training on ADNI and testing on AIBL or OASIS, the balanced accuracy was at least as high as when training and testing on AIBL or OASIS respectively, suggesting that classifiers trained on ADNI generalized well to the other datasets. In particular, training on ADNI substantially improved the classification performances on OASIS. We aimed to assess whether this was due to the larger number of subjects in ADNI. To that purpose, we performed the same experiments but with subsets of participants of equal size for each dataset. We randomly sampled populations of 70 AD patients and 70 CN participants from each of the datasets, ensuring that the demographic and clinical characteristics of the subpopulations did not differ from the original ones. As can be seen from Table 11, using the subset, the improvement disappeared for the voxel-based but remained for the regional features.

**Table 11. Influence of dataset.** Average ± SD of the balanced accuracy obtained for the reference linear SVM classifier when differentiating CN and AD subjects using voxel (4 mm smoothing) and regional (AAL2 atlas) features extracted from T1w MR images for three datasets: ADNI, AIBL and OASIS. Upper rows display results for the full population. Lower rows display results for subsets of equal size for each dataset. The subsets were obtained by randomly sampling populations of 70 AD patients and 70 CN participants from each of the datasets. Note that for the "full dataset" experiment, a balanced subset of AIBL was used (i.e. 72 CN and 72 AD subjects). When the testing dataset differs from the training dataset, there is no CV and thus no empirical SD.

|  | Training dataset | Testing dataset | Voxel-based (4 mm smoothing) | Region-based (AAL2 atlas) |
|---|---|---|---|---|
| Full dataset | ADNI | ADNI | 0.87 ± 0.025 | 0.84 ± 0.024 |
|  | AIBL | AIBL | 0.85 ± 0.003 | 0.86 ± 0.004 |
|  | ADNI | AIBL | 0.87 | 0.88 |
|  | OASIS | OASIS | 0.70 ± 0.058 | 0.71 ± 0.053 |
|  | ADNI | OASIS | 0.76 | 0.76 |
| Subset | ADNI | ADNI | 0.85 ± 0.048 | 0.81 ± 0.06 |
|  | AIBL | AIBL | 0.86 ± 0.048 | 0.85 ± 0.058 |
|  | ADNI | AIBL | 0.86 | 0.87 |
|  | OASIS | OASIS | 0.67 ± 0.063 | 0.64 ± 0.072 |
|  | ADNI | OASIS | 0.67 | 0.7 |

## 4.9. Influence of the training dataset size

Learning curves were computed to assess how the performance of linear SVM classifiers varies depending on the size of the training dataset. Using only ADNI participants, we tested four scenarios: voxel and region features extracted from T1w MRI and FDG PET images. As cross-validation, 250 iterations were run where the dataset was randomly split into a test dataset (30% of the samples) and a training dataset (70% of the samples). The maximum number of subjects used for training and testing for each of the different tasks is of 362 for CN vs AD, of 313 for CN vs pMCI and of 355 for sMCI vs pMCI. For each run, 10 classifiers were trained and evaluated on the same test set using from 10% to all of the training set (from 7% to up to 70% of the samples), increasing the number of samples used by 10% on each step. Therefore, the number of participants used for training ranged from 20 to 197 for CN, 24 to 239 for sMCI , 12 to 117 for pMCI and 17 to 166 for AD. We can observe from the learning curves in Figure 6 that, as expected, the balanced accuracy increases with the number of training samples.

Learning curves were also computed for the CN vs AD task when using larger datasets obtained by combining participants from ADNI and AIBL (balanced subset composed of 72 CN subjects and 72 AD subjects) and from ADNI, AIBL and OASIS. Results are displayed in Figure 7. We observe that for an equivalent number of subjects, combining ADNI and AIBL or only using ADNI leads to a similar balanced accuracy. For regional features, the performance is slightly higher when combining ADNI and AIBL compared to when only using ADNI, but the difference is largely within the standard deviation. The balanced accuracy keeps increasing slightly as more subjects are used for training when combining ADNI and AIBL. However, when combining ADNI, AIBL and OASIS, the performance is worse than when only using ADNI or combining ADNI and AIBL, no matter the number of subjects. This is probably due to the fact that ADNI and AIBL follow the same diagnosis and acquisition protocols, which differ from those of OASIS.

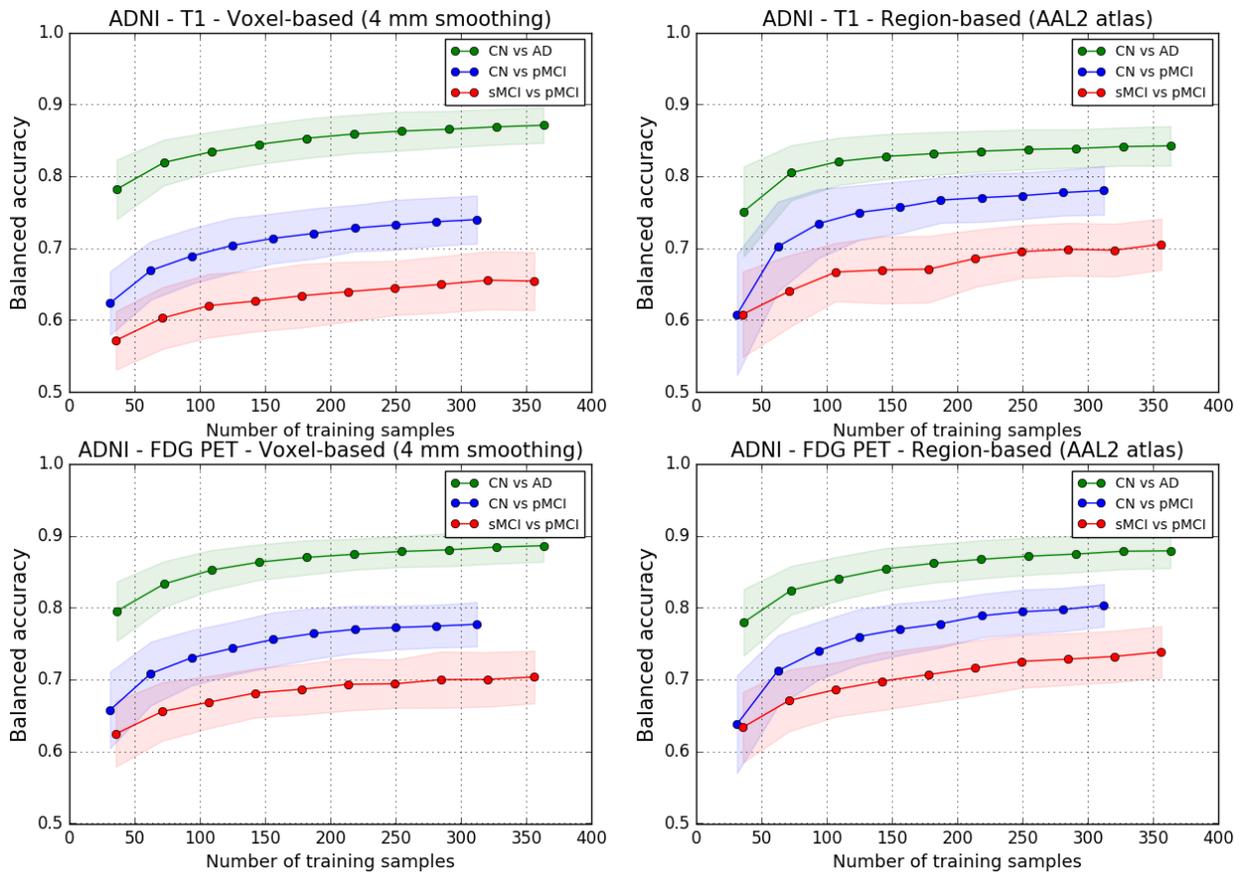

**Figure 6 Influence of training dataset size.** Learning curves for the T1w MRI (top) and FDG PET (bottom) images of ADNI participants using voxel features with 4 mm of smoothing (left) and regional features derived from the AAL2 atlas (right) for the CN vs AD, CN vs pMCI and sMCI vs pMCI tasks.

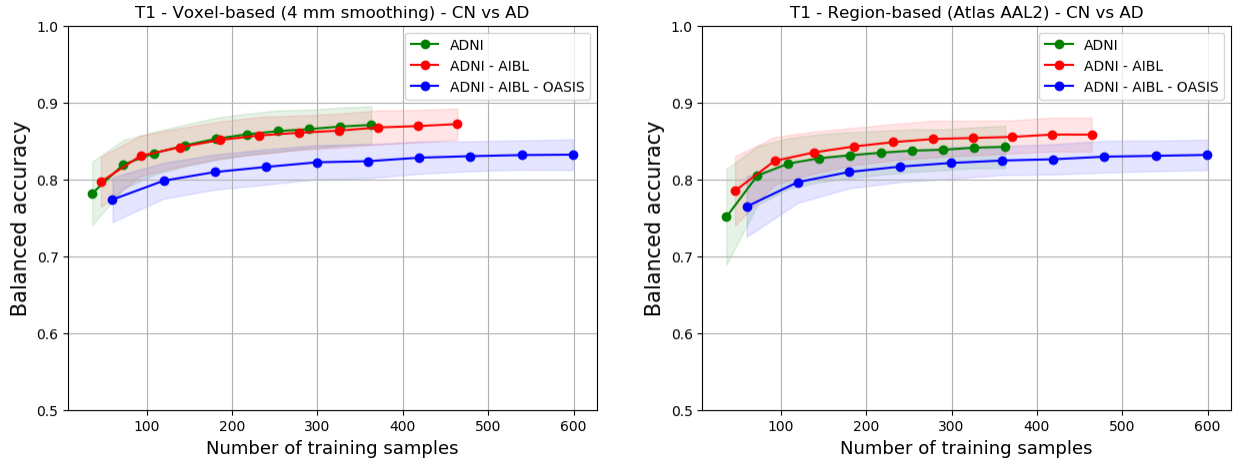

**Figure 7 Influence of training set size when combining datasets.** Learning curves for the voxel features with 4 mm of smoothing (left) and regional features derived from the AAL2 atlas (right) extracted from T1w MR images for the CN vs AD task when using subjects from ADNI only, from both ADNI and AIBL, and from ADNI, AIBL and OASIS. Note that a balanced subset of AIBL was used (i.e. 72 CN and 72 AD subjects).

## 4.10. Influence of the diagnostic criteria

We defined new classification tasks by refining the previously used diagnostic criteria using information regarding the amyloid status of each subject, when available. As can be seen in Figure 8, when comparing the performance of these tasks with their related tasks not using the amyloid status, the mean balanced accuracy is higher, or at least the same, for all the newly defined tasks. We have to note that this performance is reached in spite of counting with a lower number of subjects, given that the amyloid status is not known for all the subjects.

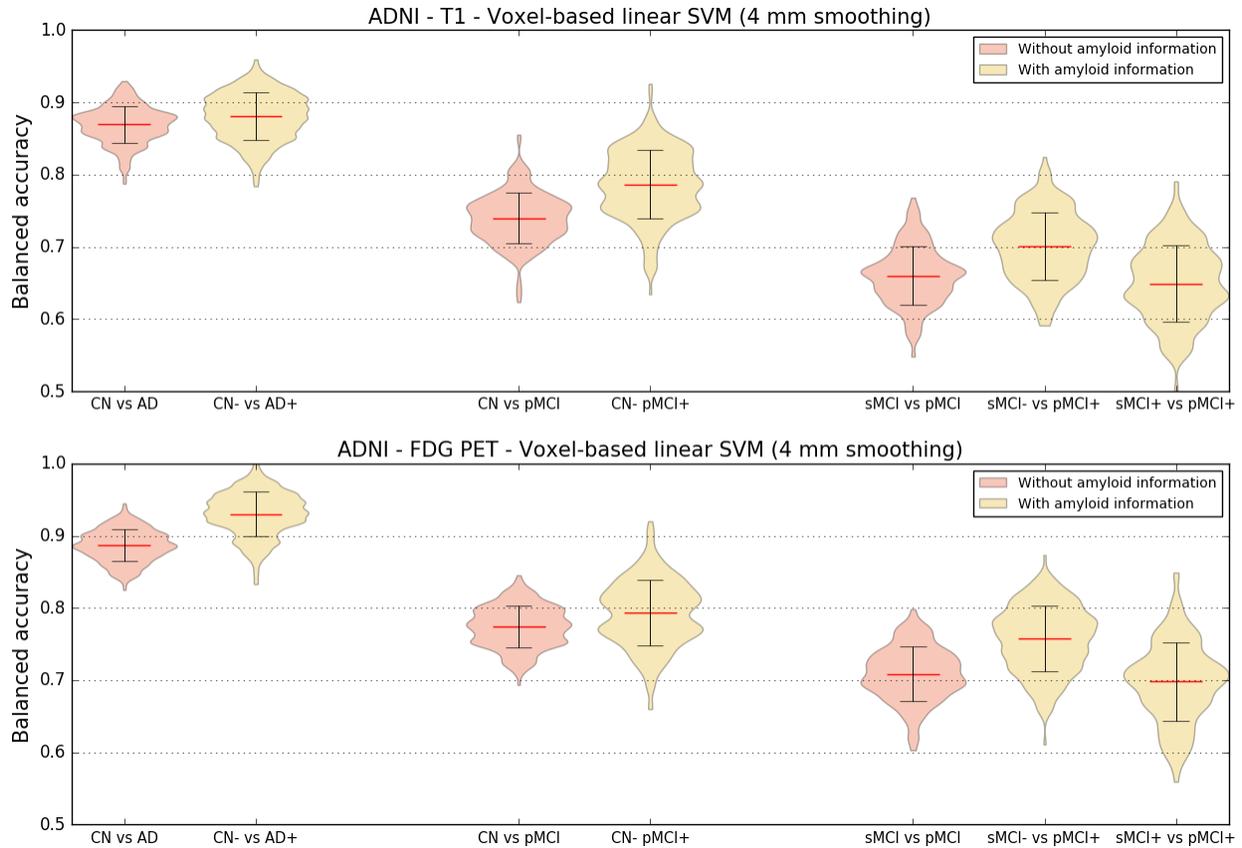

**Figure 8 Influence of diagnostic criteria.** Distribution of the balanced accuracy obtained from the T1w MRI and FDG PET images of ADNI participants using the voxel-based SVM classifier with a 4-mm smoothing for the CN- vs AD+, CN- vs pMCI+, sMCI- vs pMCI+, and sMCI+ vs pMCI+ tasks.

## 4.11. Computation time

In total, we performed 279 experiments using the SVM classifier, 155 experiments using the logistic regression classifier and 26 experiments using the random forest classifier (see Tables 6, 7 and 8 for the details of the tasks and parameters). Using a machine with 72 cores (Xeon E5-2699 @ 2.30GHz) and 256 GB of RAM, it took six days to run the 434 SVM + logistic regression experiments and eight days to run the 26 random forest experiments.

## 5. Discussion

We presented an open-source framework for the reproducible evaluation of AD classification methods that contains the following components: i) converters to normalize three publicly available datasets into BIDS; ii) standardized preprocessing and feature extraction pipelines for T1w MRI and PET; iii) standard classification algorithms; iv) cross-validation procedures following recent best practices. We demonstrated its use for the assessment of different imaging modalities, preprocessing options, features and classifiers on three public datasets.

In this work, we first aim to contribute to make evaluation of machine learning approaches in AD: i) more reproducible; ii) more objective. Reproducibility is the ability to reproduce results based on the same data and experimental procedures. Calls to increase reproducibility have been made in different fields, including neuroimaging (Poldrack et al., 2017) and machine learning (Ke et al., 2017). Reproducibility differs from replication, which is the ability to confirm results on independent data. Key elements of reproducible research include: data sharing, storing of data using community standards, fully automatic data manipulation, sharing of code. Our work can contribute to increase reproducibility of AD ML research through different aspects. A first component is the fully automatic conversion of three public datasets into the community standard BIDS. Indeed, ADNI and AIBL cannot be redistributed. Through these tools, we hope to make it easy to reproduce experiments based on these datasets without redistributing them. In particular, we offer a huge saving of time to users compared to simply making public the list of subjects used. This is particularly true for complex multimodal datasets such as ADNI (with plenty of incomplete data, multiple instances of a given modality and complex metadata). The second key component is publicly available code for preprocessing, feature extraction and classification. These contributions are gathered in Clinica[10], a freely available software platform for clinical neuroscience research studies. In addition to increased reproducibility, we hope that these tools will also make the work of researchers easier.

We also hope to contribute to more objective evaluations. Objective evaluation of a new approach (classification algorithm, feature extraction method or other) requires to test this specific component without changing the others. Our framework includes standard approaches for preprocessing and feature extraction from T1-weighted MRI and FDG PET data, and standard classification tools. These constitute a set of baseline approaches against which new methods can easily be compared. Researchers working on novel methods can then straightforwardly replace a given part of the pipeline (e.g. feature extraction, classification) with their own solution, and evaluate the added value of this specific new component over the baseline approach provided. We also propose tools for rigorous validation, largely based on recent guidelines of (Varoquaux et al., 2017) and implemented based on the standard software scikit-learn (Pedregosa et al., 2011). These include: i) large number of repeated random split to extensively assess the variability of performances; ii) reporting the full distribution of accuracies and standard deviation rather than only mean accuracies; iii) adequate nested CV for hyperparameter tuning.

We then demonstrated the use of the framework on different classification tasks based on T1 MRI and FDG PET data. Through this, we aim to provide a baseline performance to which advanced machine learning and feature extraction methods can be compared. These baseline performances are in line with the state-of-the-art results, which have been summarized in (Arbabshirani et al., 2017; Falahati et al., 2014; Rathore et al., 2017), where classification accuracies typically range from 80% to 95% for CN vs AD, and from 60% to 80% for sMCI vs pMCI. For instance, using a linear SVM, regional features (AAL2) and FDG PET data, we report 88% for CN vs AD, 80% for CN vs pMCI and 73% for sMCI vs pMCI.

---

[10] http://clinica.run

Diagnosis criteria used in ADNI are those from NINCDS-ADRDA (McKhann et al., 1984) which only rely on patients' symptoms and cognitive status. However, a definite diagnosis of AD can only be made at autopsy and clinical diagnosis has been found to be erroneous in a substantial proportion of cases (Knopman et al., 2001). In the past decade, substantial progress has been made in the diagnosis of AD. In particular, it has been suggested to not only rely on clinical and cognitive evaluations but also on imaging and cerebrospinal fluid (CSF) biomarkers. This has resulted in new diagnostic criteria. Even though the gold-standard remains postmortem examination, this has led to a more accurate diagnosis of AD during the life of the patient. In particular, the presence of beta-amyloid and/or tau proteins has been proposed in IWG (Dubois et al., 2007), IWG-2 (Dubois et al., 2014) and NIA-AA (Albert et al., 2011; Jack et al., 2011; McKhann et al., 2011; Sperling et al., 2011). In this work, we assessed if using amyloid-refined diagnosis groups improved the performance. Amyloid status was determined from each participant's amyloid PET scan (PiB or AV45). We found that classification using amyloid-refined diagnoses always performed better or at least similarly to the related tasks using NINCDS-ADRDA diagnoses, even though the training sets then comprise fewer individuals.

Classifications from FDG PET consistently performed better across tasks, features and classification methods than from T1w MRI. Some studies support our finding (Dukart et al., 2011, 2011; Gray et al., 2013; Ota et al., 2015; Young et al., 2013) while others do not find a difference in performance (Hinrichs et al., 2009; Zhang et al., 2011; Zhu et al., 2014). Given the larger sample size of our study and the rigorous evaluation design, we believe that the superior performance of FDG PET compared to MRI is a robust finding. It is likely due to the fact that hypometabolism can be detected earlier in the disease course, before atrophy (Jack et al., 2010b).

Diverse parameters and options are used as for preprocessing and feature extraction in AD machine learning studies. Their influence on classification performance is not clear and constitutes a problem for the comparability of classification methods. We assessed the effect of the choice of atlas, of degree of smoothing, of the correction of PET images for partial volume effect, and of the type of features (regions or voxels). We found no systematic effect of each of these different components on the performances. Some studies found an influence of the atlas on the classification performance (Ota et al., 2015, 2014). However, the number of subjects in this study was small. In (Chu et al., 2012), an improvement of 3% was found when using a combination of a few ROIs compared to using all the voxels. In our study, a much larger number of subjects and a strict validation process were used.

We compared three widely used classification methods: SVM, logistic regression with L2 regularization and random forests. Our main finding was the underperformance of the latter. This might be caused by the nature of brain imaging data that contains relatively homogeneous values, and which should show dependence across voxels or brain regions. These characteristics of the data could explain why techniques trying to find a smooth combination of features, such as those using L2 regularization, are more suited for single modality classification problem. On the other hand, random forests or other ensemble methods could be useful when combining features from different modalities such as images, clinical data and cognitive scores, as done in (Moradi et al., 2015; Sørensen et al., 2018). In other papers comparing several standard classification algorithms such as SVM, LDA or Naive Bayes (Aguilar et al., 2013; Cabral et al., 2015; Sabuncu et al., 2015), results did not show differences between methods.

We also assessed the influence of class imbalance, which in our datasets ranges from very mild (1.2 times more CN than AD for ADNI) to moderate (1.7 times more CN than pMCI and 2 times more sMCI than pMCI for ADNI) to very strong (6.1 times more CN than AD in AIBL). In the case of voxel-based features, we found that a very strong class imbalance (as in the case of AIBL where the proportion is 6 to 1) leads to lower performance but that moderate class imbalance (up to 2 to 1 in ADNI) are adequately handled. On the other hand, there was no influence of class imbalance for regional features. This highlights that it may be beneficial to use balanced groups for training when there is a very strong class imbalance and when using very high dimensional features.

We assessed the influence of various components on classification performance: modality (T1w MRI vs PET), type of features, choice of atlas, PVC, smoothing, classifier. Other studies have assessed the influence of other components: different types of anatomical features including volume, cortical thickness and other surface characteristics (Gómez-Sancho et al., 2018; Schwarz et al., 2016; Westman et al., 2013), feature selection techniques (Tohka et al., 2016), normalization to intracranial volume (Voevodskaya et al., 2014; Westman et al., 2013). Moreover, (Tohka et al., 2016) compared LASSO and elastic-net to SVM and found that the former methods provide increased performance. Assessing the influence of these different components could also be done using our framework. In this paper, we restricted the application of the framework to a set of components that were chosen for the following reasons. Voxel-based and regional features were both included because they are widely used. On the other hand, cortical measures based on Freesurfer were not included due to their computational cost. PVC is a very common preprocessing for PET data. Smoothing is widely used for voxel-based analyses in the neuroimaging community and it seemed useful to assess its influence. Nevertheless, there is always some arbitrariness in such choices and it would be interesting to study other components with the framework.

In this work, we used predefined features (at the region or voxel-level). Another family of approaches that should be mentioned is that of methods that learn features directly from the data. Patch-based methods aim to automatically learn the nonlocal similarity between a subject and a training set (Coupé et al., 2015, 2012). Also, deep learning approaches can automatically learn relevant features at multiple scales, and have recently become popular for automatic classification of AD (Bäckström et al., 2018; Liu et al., 2018; Lu et al., 2018; Suk et al., 2017). Both types of approaches have led to promising results (e.g. from 73% to 83% for pMCI vs sMCI). Moreover, various works have proposed to use different types of data-driven feature selection (e.g. univariate statistical tests, multivariate approaches) (Chu et al., 2012; Tohka et al., 2016; Vemuri et al., 2008) and dimensionality reduction (e.g. principal component analysis, manifold learning) (Beheshti et al., 2015; Guerrero et al., 2014; Liu et al., 2015; Salvatore et al., 2015). These approaches have the potential to improve the performance but they need to be validated using rigorous cross-validation procedures (Eskildsen et al., 2013; Maggipinto et al., 2017). The evaluation of the added value of all these approaches could be done using our framework. This is out of the scope of the present paper and is left for future work.

Using multiple datasets is important to assess if the performances are robust to different populations, acquired in different conditions. A first component consisted in performing the same experiments on different datasets. We found that classification results were similar for ADNI and AIBL datasets, but much lower for OASIS. The lower performance for OASIS is likely due to the diagnosis criteria which are less rigorous (in OASIS, all participants with CDR>0 are considered AD). It is also valuable to know how a classifier will perform when trained on one dataset and tested on another one. The classifiers trained on ADNI data generalized well to AIBL and OASIS. Interestingly, for OASIS, the performances were substantially increased when training on ADNI compared to when training on OASIS. Such improvement may arise from several factors: larger training set size, higher image quality or stricter diagnostic criteria. When using subsets of equal size, the improvement obtained for voxel-based features disappeared, suggesting that increased training set size is important, in particular when using very high dimensional features. On the other hand, for regional features, training on the ADNI subset improved performances compared to training on the OASIS subset, suggesting that other factors (image quality, stricter diagnostic criteria) contribute to the improvement. In general, we can say that classifiers are able to generalize across different datasets, as is also concluded in (Dukart et al., 2013; Sabuncu et al., 2015) particularly if they are obtained using large multicentric datasets with strict diagnostic criteria, as is the case for ADNI.

Unsurprisingly, increased training set size led to increased classification performances. This improvement of the results depending on the training set size has also been found in other studies such as (Abdulkadir et al., 2011; Chu et al., 2012; Franke et al., 2010). One can note that when combining multiple datasets, performances also increased with training set size. However, when combining OASIS together with ADNI and AIBL, the performance was lower than when using only AIBL and ADNI. This is consistent with the fact that performances for OASIS are systematically lower than those obtained on ADNI and AIBL. Again, this is likely due to diagnostic criteria which are less rigorous in OASIS. Interestingly, with the current number of samples available, the point where the results stop improving has not been reached. The performance of the classifier reaches a limit imposed by the number of images that have been provided for training, meaning that more data are necessary to find the top performance of a classifier. These results highlight the need for more publicly available datasets, on which most of the current research in the field relies.

## 6. Conclusions

Our framework for reproducible classification experiments aims to address current issues faced in the area of machine learning-based AD classification, such as comparability and reproducibility of the results. Its application to T1w MRI and FDG PET data allowed the extensive assessment of the influence of imaging modality, preprocessing options, features and algorithms on the performances. These results provide a baseline performance against which other approaches can be compared. We hope that both the framework and the experimental results will be useful to researchers working in the AD field.

# Acknowledgments

The research leading to these results has received funding from the program "Investissements d'avenir" ANR-10-IAIHU-06 (Agence Nationale de la Recherche-10-IA Agence Institut Hospitalo-Universitaire-6), ANR-11-IDEX-004 (Agence Nationale de la Recherche-11- Initiative d'Excellence-004, project LearnPETMR number SU-16-R-EMR-16), from the European Union H2020 program (project EuroPOND, grant number 666992, project HBP SGA1 grant number 720270), from the ICM Big Brain Theory Program (project DYNAMO), from the European Research Council (to Dr Durrleman project LEASP, grant number 678304), from the joint NSF/NIH/ANR program "Collaborative Research in Computational Neuroscience" (project HIPLAY7, grant number ANR-16-NEUC-0001-01) and from the "Contrat d'Interface Local" program (to Dr Colliot) from Assistance Publique-Hôpitaux de Paris (AP-HP). N.B. receives funding from the People Programme (Marie Curie Actions) of the European Union's Seventh Framework Programme (FP7/2007-2013) under REA grant agreement no. PCOFUND-GA-2013-609102, through the PRESTIGE programme coordinated by Campus France.

Data collection and sharing for this project was funded by the Alzheimer's Disease Neuroimaging Initiative (ADNI) (National Institutes of Health Grant U01 AG024904) and DOD ADNI (Department of Defense award number W81XWH-12-2-0012). ADNI is funded by the National Institute on Aging, the National Institute of Biomedical Imaging and Bioengineering, and through generous contributions from the following: AbbVie, Alzheimer's Association; Alzheimer's Drug Discovery Foundation; Araclon Biotech; BioClinica, Inc.; Biogen; Bristol-Myers Squibb Company; CereSpir, Inc.; Cogstate; Eisai Inc.; Elan Pharmaceuticals, Inc.; Eli Lilly and Company; EuroImmun; F. Hoffmann-La Roche Ltd and its affiliated company Genentech, Inc.; Fujirebio; GE Healthcare; IXICO Ltd.; Janssen Alzheimer Immunotherapy Research & Development, LLC.; Johnson & Johnson Pharmaceutical Research & Development LLC.; Lumosity; Lundbeck; Merck & Co., Inc.; Meso Scale Diagnostics, LLC.; NeuroRx Research; Neurotrack Technologies; Novartis Pharmaceuticals Corporation; Pfizer Inc.; Piramal Imaging; Servier; Takeda Pharmaceutical Company; and Transition Therapeutics. The Canadian Institutes of Health Research is providing funds to support ADNI clinical sites in Canada. Private sector contributions are facilitated by the Foundation for the National Institutes of Health (www.fnih.org). The grantee organization is the Northern California Institute for Research and Education, and the study is coordinated by the Alzheimer's Therapeutic Research Institute at the University of Southern California. ADNI data are disseminated by the Laboratory for Neuro Imaging at the University of Southern California.

The OASIS project was supported by the following grants: P50 AG05681, P01 AG03991, R01 AG021910, P20 MH071616, and U24 RR021382.